\documentclass[conference]{IEEEtran}
\let\labelindent\relax
\usepackage[inline]{enumitem}
\usepackage{subcaption}
\usepackage{epsfig}
\usepackage{dblfloatfix}
\usepackage{float}
\usepackage{hyperref} 

\IEEEoverridecommandlockouts                              

\IEEEpubid{\makebox[\columnwidth]{978-1-5386-5541-2/18/\$31.00~\copyright2020 IEEE \hfill} \hspace{\columnsep}\makebox[\columnwidth]{ }}

\title{\LARGE \bf
From Simulation to Real World Maneuver Execution using Deep Reinforcement Learning
}

\author{Alessandro Paolo Capasso$^{1}$, Giulio Bacchiani$^{1}$ and Alberto Broggi$^{2}$
\thanks{$^{1}$Alessandro Paolo Capasso is with VisLab - University of Parma, Parma, Italy {\tt\small alessandro.capasso@unipr.it}}%
\thanks{$^{1}$Giulio Bacchiani and $^{2}$Alberto Broggi are with VisLab srl, an Ambarella Inc. company - Parma, Italy {\tt\small gbacchiani@ambarella.com}, {\tt\small broggi@vislab.it}}%
}

\begin{document}

\maketitle
\IEEEpubidadjcol

\begin{abstract}

Deep Reinforcement Learning has proved to be able to solve many control tasks in different fields, but the behavior of these systems is not always as expected when deployed in real-world scenarios. This is mainly due to the lack of domain adaptation between simulated and real-world data together with the absence of distinction between train and test datasets. In this work, we investigate these problems in the autonomous driving field, especially for a maneuver planning module for roundabout insertions. In particular, we present a system based on multiple environments in which agents are trained simultaneously, evaluating the behavior of the model in different scenarios. Finally, we analyze techniques aimed at reducing the gap between simulated and real-world data showing that this increased the generalization capabilities of the system both on unseen and real-world scenarios. 
\end{abstract}

\section{INTRODUCTION}

Deep Reinforcement Learning (DRL)~\cite{drl} has made impressive achievements in solving complex tasks both in discrete control-space problems like Go~\cite{go} and Atari~\cite{atari} games, and in continuous ones~\cite{continuous}. However, many approaches do not consider to split train and test environments for evaluating the performance of the system, increasing the risk of overfitting on specific scenarios. In these cases, the system becomes specialized to solve difficult tasks in the training environments but often fails when deployed on unseen ones. In fact, the lack of generalization makes the system unusable in real-world scenarios where uncertainty and noise elements, as well as novel situations, will surely be encountered~\cite{uncert}.


In this work we focus on the problem of overfitting for DRL agents trained to execute autonomous driving maneuverssuch as roundabout insertions. We develop a training pipeline inspired by the typical supervised learning approach featuring distincted train, validation and test datasets. Indeed, agents are trained in four \textit{training} environments (Fig.~\ref{fig:synthetic_roundabouts}) simultaneously, using an additional \textit{validation} scenario (Fig.~\ref{fig:synthetic_stadio}) to obtain the best model based on the results achieved on this scenario and finally measuring the generalization performance on a further \textit{test} environment (Fig.~\ref{fig:synthetic_campus}), which do not have any active role during the training phase. We proved that the use of multiple training environments together with the validation scenario is essential to avoid overfitting for the proposed task. Moreover, even if we are focusing on the specific maneuver of roundabout insertions, the techniques proposed in this work can be applied to different tasks such as intersection handlings or overtakings.

Since Deep Reinforcement Learning algorithms generally require a large amount of experiences for learning a task, the majority of DRL trainings is done in simulation and the transfer of knowledge from synthetic to real-world data could be hard to achieve as explained in~\cite{gap}. In realistic graphic simulators like CARLA~\cite{carla} and GTA-based platforms~(\cite{gta1},~\cite{gta2}), the transfer of a policy between synthetic and real domains is still an open problem even if recent works have shown encouraging results~(\cite{tp1},~\cite{tp2}). In contrast, our approach consists in the development of synthetic representations of real scenarios (Fig.~\ref{fig:real_roundabouts}) built with Cairo graphic library~\cite{cairo} in order to reduce the initial gap between simulation and real domains; however, variabilities in the human behaviors and noise introduced by detection systems, make simulation and reality still differ: for this reason we introduce source of uncertainties in order to make the agent more robust and thus suitable to be deployed in real world scenarios.

Finally, we compare the behavior of our system with the one developed in~\cite{iri} showing that our model achieves better results both on unseen and real-world scenarios.

\section{RELATED WORK}

\begin{figure*}
  \vspace{0.3cm}
  \centering
  \begin{subfigure}{.19\linewidth}
    \centering
    \includegraphics[width =\linewidth]{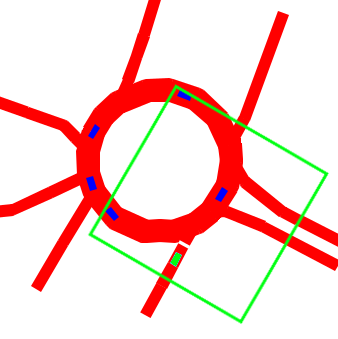}    
    \caption{Training roundabout 1}
    \label{fig:synth_barriera}
  \end{subfigure}
  \hspace{0.5cm}
  \begin{subfigure}{.19\linewidth}
    \centering
    \includegraphics[width =\linewidth]{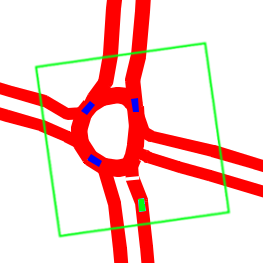}
    \caption{Training roundabout 2}
    \label{fig:synth_four}
  \end{subfigure}
  \hspace{0.5cm}
  \begin{subfigure}{.19\linewidth}
    \centering
    \includegraphics[width =\linewidth]{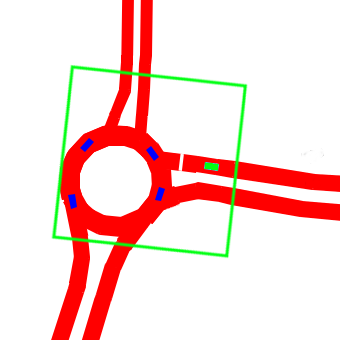}
    \caption{Training roundabout 3}
    \label{fig:synth_parmamia}
  \end{subfigure}
  \hspace{0.5cm}
  \begin{subfigure}{.19\linewidth}
    \centering
    \includegraphics[width =\linewidth]{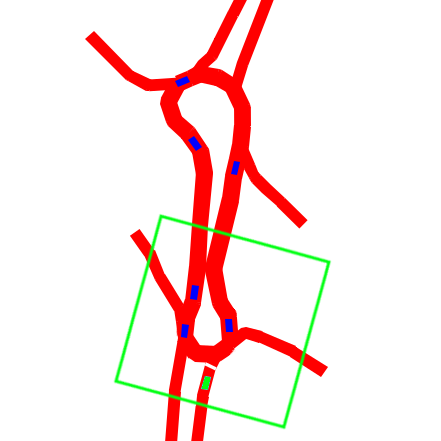}
    \caption{Training roundabout 4}
    \label{fig:synth_pear}
  \end{subfigure}  
  \caption{Synthetic representations of real roundabouts. The green squares highlights the surroundings perceived by the green agent, while the traffic inside the roundabout is composed by blue vehicles.}
  \label{fig:synthetic_roundabouts}
\end{figure*}

\begin{figure*}
  \centering
  \begin{subfigure}{.18\linewidth}
    \centering
    \includegraphics[width =\linewidth]{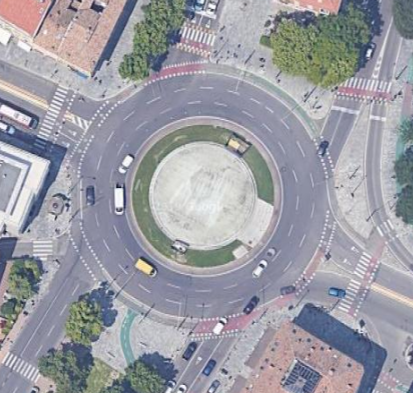}    
    \caption{Real roundabout 1}
    \label{fig:real_barriera}
  \end{subfigure}
  \hspace{0.5cm}
  \begin{subfigure}{.184\linewidth}
    \centering
    \includegraphics[width =\linewidth]{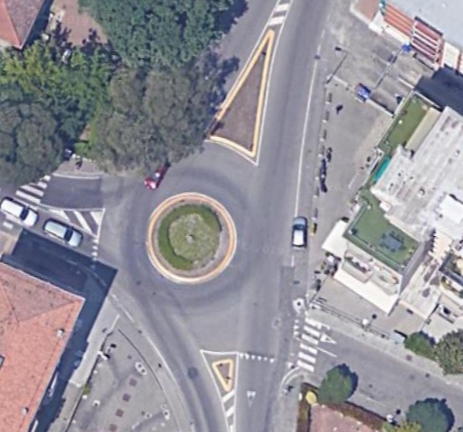}
    \caption{Real roundabout 2}
    \label{fig:real_four}
  \end{subfigure}
  \hspace{0.5cm}
  \begin{subfigure}{.183\linewidth}
    \centering
    \includegraphics[width =\linewidth]{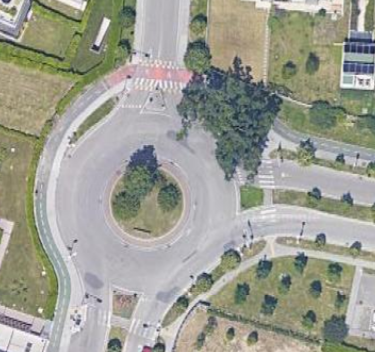}
    \caption{Real roundabout 3}
    \label{fig:real_parmamia}
  \end{subfigure}
  \hspace{0.5cm}
  \begin{subfigure}{.206\linewidth}
    \centering
    \includegraphics[width =\linewidth]{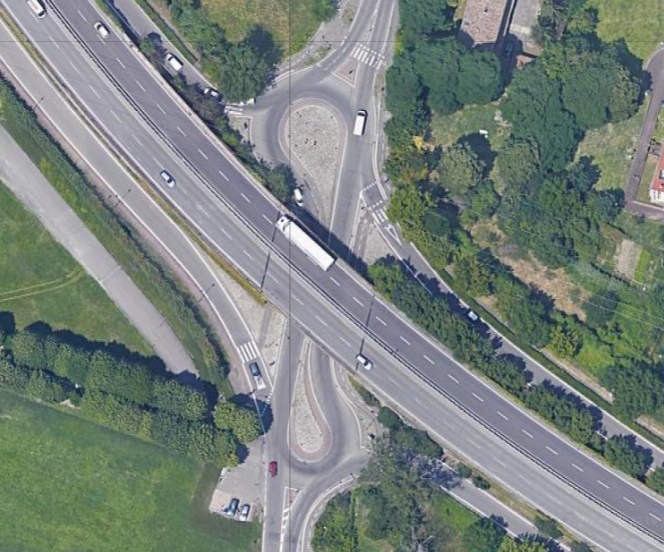}
    \caption{Real roundabout 4}
    \label{fig:real_pear}
  \end{subfigure}  
  \caption{Top view of real roundabouts of the city of Parma used to generate the synthetic ones (Fig.~\ref{fig:synthetic_roundabouts}).}
  \label{fig:real_roundabouts}
\end{figure*}

\begin{figure*}[!b]
  \centering
  \begin{subfigure}{.19\linewidth}
    \centering
    \includegraphics[width =\linewidth]{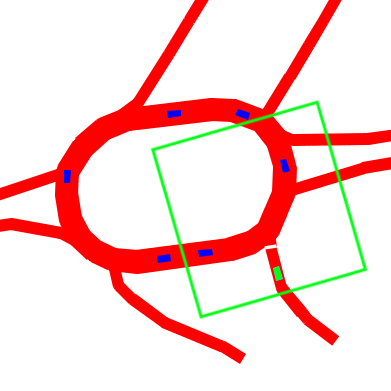}    
    \caption{Validation roundabout}
    \label{fig:synthetic_stadio}
  \end{subfigure}
  \hspace{0.5cm}
  \begin{subfigure}{.213\linewidth}
    \centering
    \includegraphics[width =\linewidth]{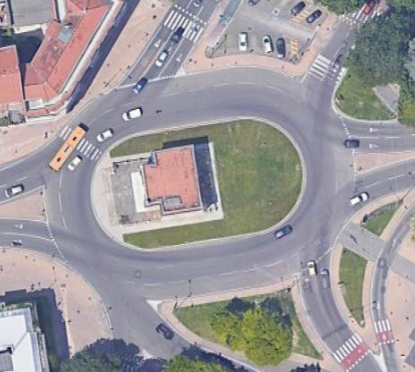}
    \caption{Real roundabout}
    \label{fig:real_stadio}
  \end{subfigure}
  \hspace{0.5cm}
    \begin{subfigure}{.19\linewidth}
    \centering
    \includegraphics[width =\linewidth]{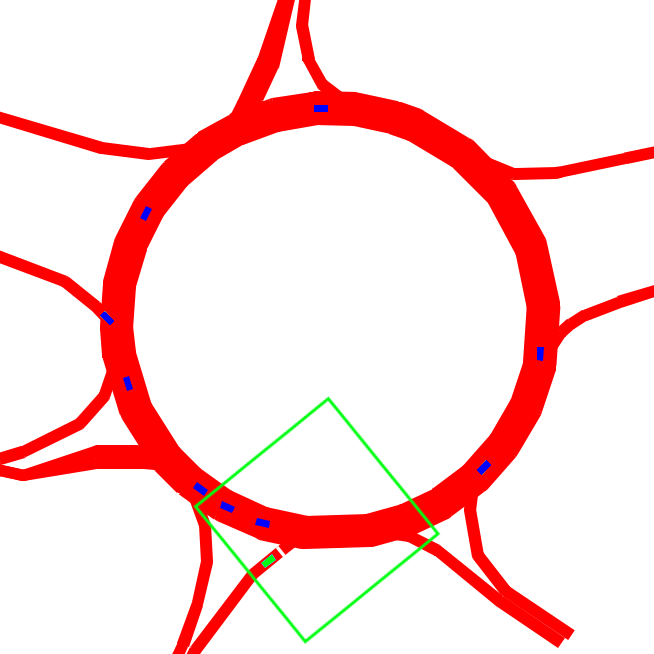}    
    \caption{Test roundabout}
    \label{fig:synthetic_campus}
  \end{subfigure}
  \hspace{0.5cm}
  \begin{subfigure}{.19\linewidth}
    \centering
    \includegraphics[width =\linewidth]{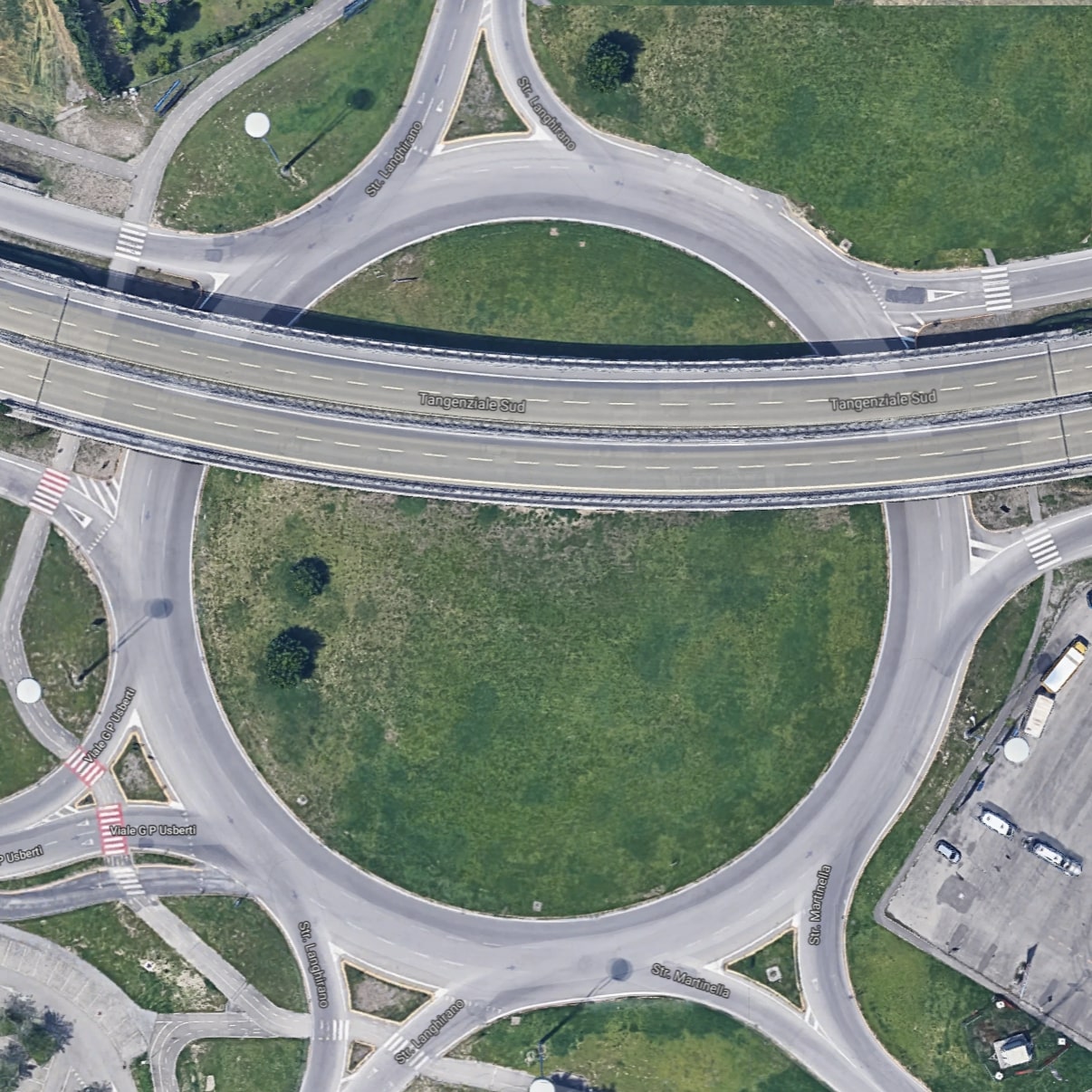}
    \caption{Real roundabout}
    \label{fig:real_campus}
  \end{subfigure}
  \caption{Top view of real roundabouts ((b), (d)) used to generate validation and test synthetic environments ((a), (c)).}
  \label{fig:val_test_roundabouts}
\end{figure*}

Many recent works separate train and test scenarios creating levels with different difficulties, with the aim of helping the training phase by gradually making the problem more challenging. In~\cite{cobbe} the problem of generalization in DRL has been addressed showing that increasing the number of training levels, the risk of overfitting decreases. A similar approach is followed in~\cite{zhang} and ~\cite{justesen}, where levels have been generated using gridworld mazes and the General Video Game AI framework (GVG-AI)~\cite{gvg} respectively; both of these approaches show that agents have a high capacity of memorizing specific levels of a training set but the performances on test scenarios could be very different based on the complexity of levels or the training set size. Moreover, in~\cite{zhang} it has also been proved that techniques like random starts~(\cite{nair},~\cite{haus}) and sticky actions~\cite{machado} are often unuseful to avoid overfitting. 

Furthermore, achieving generalization in unseen simulated environments cannot guarantee a correct behavior in real-world scenarios due to its unpredictable elements~(\cite{uncert},~\cite{gap}). In~\cite{packer}, authors evaluated the performances of agents changing some environmental parameters such as robot mass and push force magnitude in CartPole and MountainCar~(\cite{cartpole},~\cite{cartpole2}) or the moment of inertia in Acrobot~\cite{acrobot}.

Our approach differs from these works since we do not generate thousands of levels, but we train the system on a fixed number of \textit{training} environments (Fig.~\ref{fig:synthetic_roundabouts}) with distinct lengths and shapes. During the training process we evaluate the performance of the system on the \textit{validation} roundabout (Fig.~\ref{fig:synthetic_stadio}), and then tested on the \textit{test} roundabout (Fig.~\ref{fig:synthetic_campus}), that is the same used in~\cite{iri} to quantify generalization performances. Comparing our results with those obtained in~\cite{iri}, we show that our model achieves better generalization performances on unseen environments since the lack of train-test distinction and a proper amount of scenarios in~\cite{iri}, lead the system to overfit on the training environment (Fig.~\ref{fig:synth_parmamia}).

\section{NOTATION}

\subsection{Reinforcement Learning}
In Reinforcement Learning (RL)~\cite{rl} an agent interacts with an environment trying to learn the best policy by trials and errors. The RL problem can be described as a Markov Decision Process (MDP) $M = (S, A, P, r, P_0)$, with the state space $S$, the action space $A$, the state transition probability $P(s_{t+1}|s_t, a_t)$, the reward function $r$ and finally the probability distribution $P_0$ of the initial state $s_0 \in S$. The goal of the RL agent is to learn a policy that maximizes the \textit{expected return}; at time $t$ it is generally defined as: $R_t = \sum_t^T r_t + \gamma r_{t+1} + \cdots + \gamma^{T-t}r_{T}$, where $T$ is the terminal time step and $\gamma$ is the discount factor. 

In our environments two types of vehicles interact among each other: the \textit{passives} and the \textit{actives}. We will give further details on agents in the following sections.

\begin{figure}[h]
  \vspace{0.3cm}
  \centering
  \begin{subfigure}{.4\linewidth}
    \centering
    \includegraphics[width =\linewidth]{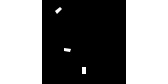}    
    \caption{Obstacles}
    \label{fig:obstacles}
  \end{subfigure}
  \hspace{-1.0cm}
  \begin{subfigure}{.4\linewidth}
    \centering
    \includegraphics[width =\linewidth]{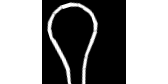}
    \caption{Path}
    \label{fig:path}
  \end{subfigure}
  \\
  \vspace{0.25cm}
  \begin{subfigure}{.4\linewidth}
    \centering
    \includegraphics[width =\linewidth]{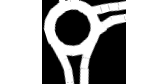}    
    \caption{Navigable space}
    \label{fig:nav_space}
  \end{subfigure}
  \hspace{-1.0cm}
  \vspace{0.25cm}
  \begin{subfigure}{.4\linewidth}
    \centering
    \includegraphics[width =\linewidth]{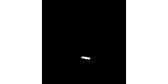}
    \caption{Stop line}
    \label{fig:stop_line}
  \end{subfigure}  
  \caption{Example of a single visual input of the \textit{Training roundabout 3} (Fig.~\ref{fig:synth_parmamia}) perceived by the agent.}
  \label{fig:semantic_layers}
\end{figure}

\subsection{Passive Agents}
\label{sec:passive_agents}

The traffic inside the roundabout is composed by \textit{passive} agents which are trained by \textit{Multi-agent A3C}~\cite{mts}; during the training phase several learning agents share the same environment instance so that actions of an agent may potentially affect the state of other agents and vice versa. In this way they learn to cooperate and negotiate with each other in order to perform the best actions in terms of time and safety. The network architecture, as well as the reward shaping, is the same used in~\cite{mts}; it takes a sequence of four visual inputs, each one composed by three images 84x84 (obstacles, path and navigable space) (Fig.~\ref{fig:obstacles},~\ref{fig:path},~\ref{fig:nav_space}) corresponding to 50x50 meters of the agent's sourrounding. Along with this visual input the net receives a non-visual sensory channel composed by 
\begin {enumerate*} [label=\itshape\alph*\upshape)]
\item the agent speed, \item the target speed (the velocity that the agent should not exceed), \item the aggressiveness that modules the driving behavior of the vehicle and \item the distance to the goal.
\end {enumerate*} 
The outputs of the network are the state-value estimation and the probabilities of three possible actions: accelerate ($1 \frac{m}{s^2}$), keep the same speed and brake ($-2 \frac{m}{s^2}$).

In this work, we train \textit{passive} agents in all the roundabouts illustrated in Fig.~\ref{fig:synthetic_roundabouts}, Fig.~\ref{fig:synthetic_stadio} and Fig.~\ref{fig:synthetic_campus} in a \textit{Multi-environment Multi-agent} fashion. We introduce a normalization factor in the reward since the paths that vehicles should follow have different length because of various shapes and sizes of the training roundabouts. However, since we are investigating the generalization performance for the insertion maneuver (performed by the \textit{active} vehicle) we want that \textit{passive} agents behave as good as possible in each roundabout and for this reason we train them on all the environments. Anyway, the techniques proposed in this work to increase generalization for \textit{active} agents, could be applied to \textit{passives} without additional efforts.

\subsection{Active Agents}

\textit{Active} vehicles perform the insertion maneuver in the roundabout and they are trained through \textit{Delayed A3C} (D-A3C)~\cite{iri} since it has been demonstrated that the performances of this algorithm overcome those achieved by the classic \textit{A3C}~\cite{a3c} in this setting. In \textit{D-A3C} the system collects the contributions of each asynchronous learner during the whole episode, sending the updates to the global network only at the end of the episode, while in the classic A3C this exchange is performed at fixed time intervals.

The network architecture and the reward function are the same used in~\cite{iri}. Agents receive a sequence of four visual inputs, each one composed by four images 84x84 (obstacles, path, navigable space and stop line) (Fig.~\ref{fig:semantic_layers}), corresponding to a square area of 50x50 meters around the agent, and a non-visual input sensory channel consisting of four entities:
\begin {enumerate*} [label=\itshape\alph*\upshape)]
\item the agent speed, \item the target speed, \item the aggressiveness in the maneuver execution and \item the last action performed by the vehicle.
\end {enumerate*}
The outputs of the module are the state-value estimations and the probabilities of three possible states:
\begin{itemize}
\item Permitted: the entry area is predicted as free by the agent and it produces an acceleration unless the target speed is reached.
\item Not Permitted: the agent predicts the entry area of the roundabout as busy and it produces a deceleration such that comfort constraints are followed.
\item Caution: the agent predicts the entry area as not completely free and the vehicle approaches the roundabout with prudence.
\end{itemize}
Further details of the output and the reward function are given in~\cite{iri}.

\section{GENERALIZATION TECHNIQUES}
\label{sec:techniques}

The following techniques are designed to avoid overfitting and therefore they are used during the training phase of \textit{active} vehicles.

\subsection{Multi-environment System}
\label{sec:multi-env_system}

The \textit{Multi-environment System} consists in four \textit{training} roundabouts (Fig.~\ref{fig:synthetic_roundabouts}), in which \textit{active} agents are trained simultaneously, and a further scenario used as \textit{validation} environment (Fig.~\ref{fig:synthetic_stadio}). For every roundabout we create as many instances as the number of entry lanes such that in every instance only one active agent performs the immission maneuver from a specific entry. During the training phase, we evaluate the generalization performance of the system on the \textit{validation} roundabout (Fig.~\ref{fig:synthetic_stadio}) in order to save the best model based on the results obtained on this scenario: \textit{active} vehicles acting in this environment do not compute or send back the gradient but they perform the maneuver with the latest network weights pulled at the beginning of each episode. 

The number of \textit{passive} agents populating each instance of training and validation roundabouts allows the \textit{active} agent to handle different traffic conditions (Table~\ref{max_passive}), such that the agent could observe different states of the environment.

\begin{table}[h] \centering
\caption{Maximum number of \textit{passive} vehicles popoulating each instance of training (Fig.~\ref{fig:synthetic_roundabouts}) and validation (Fig.~\ref{fig:synthetic_stadio}) roundabouts.}
\label{max_passive}
\begin{center}
\begin{tabular}{|c|c|}
\hline
\textbf{Roundabout Instance} & \textbf{\#Passives}\\
\hline
Training roundabout 1 & 6\\
\hline
Training roundabout 2 & 3\\
\hline
Training roundabout 3 & 6\\
\hline
Training roundabout 4 & 6\\
\hline
Validation roundabout & 9\\
\hline
\end{tabular}
\end{center}
\end{table}

For computational reasons we chose five scenarios during the training phase for a total amount of 19 \textit{active} agents performing the insertion maneuver (one for each entry lane) interacting with 114 \textit{passive} vehicles, namely the sum of the maximum number of \textit{passives} used for every single instance of each environment~(Table~\ref{max_passive}) multiplied for the number of entry lanes of each roundabout. 


\begin{figure}[h]
  \vspace{0.3cm}
  \centering
  \hspace{-1cm}
  \begin{subfigure}{.45\linewidth}
    \centering
    \includegraphics[width =\linewidth]{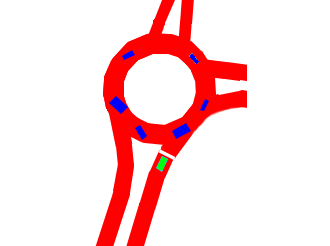}    
    \caption{Perception Noise}
    \label{fig:perception_noise}
  \end{subfigure}
  \hspace{0.5cm}
  \begin{subfigure}{.356\linewidth}
    \centering
    \includegraphics[width =\linewidth]{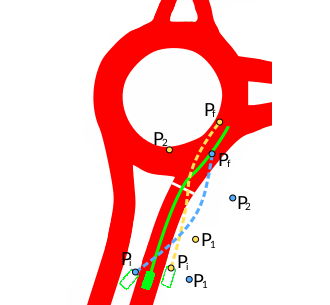}    
    \caption{Localization Noise}
    \label{fig:localization_noise}
  \end{subfigure}  
  \caption{Examples of noise injected in the position, shape and size of \textit{passive} agents~(a) and in the path of the \textit{active} agent~(b).}
  \label{fig:noise_passives_path}
\end{figure}

\begin{figure}[h]
  \centering
  \begin{subfigure}{.45\linewidth}
    \centering
    \includegraphics[width =\linewidth]{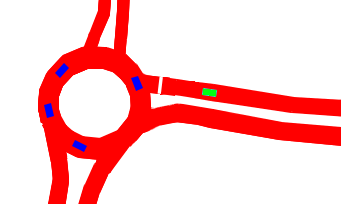}    
    \caption{Original}
    \label{fig:original_env}
  \end{subfigure}
  \begin{subfigure}{.45\linewidth}
    \centering
    \includegraphics[width =\linewidth]{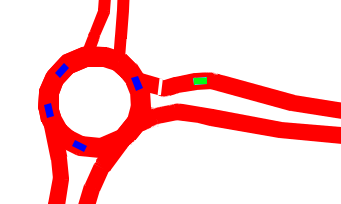}    
    \caption{Noised}
    \label{fig:noised_env}
  \end{subfigure}  
  \caption{Example of a noised entry lane (b) generated from the original shape (a) of the roundabout in Fig.~\ref{fig:synth_parmamia}.}
  \label{fig:noise_envs}
\end{figure}

\subsection{Noise Injection}
\label{sec:noise_inj}

Considering the simulated environments in which agents are trained, a domain adaptation between synthetic and real data should be performed in order to make the system usable in real world. Indeed, data received from perception and localization systems on board of a self-driving vehicle could be very different from those using during the simulation. Moreover, additional noise elements will surely be encountered given the stochastic nature of real world scenarios~\cite{uncert}. 

To reproduce this unpredictable noise in the simulator, we introduce the following artificial elements:
\begin{itemize}
\item \textit{Perception Noise}: it consists in two different uncertainty sources;
firstly, in the position ($x, y$ coordinates), size (width and height) and heading of the \textit{passive} agents perceived from the \textit{active} one as shown in Fig.~\ref{fig:perception_noise}; secondly, we introduce errors in the detection of \textit{passive} vehicles: each time step a \textit{passive} agent could not be detected by the \textit{active} one, such that the four sequential images of obstacles (Fig.~\ref{fig:obstacles}) given as input to the neural network may contain errors.

\item \textit{Localization Noise}: as in~\cite{iri}, we generate a new path of the \textit{active} vehicle perturbing the original one using Cubic B\'ezier curves~\cite{bezier} (Fig.~\ref{fig:localization_noise}). This error is useful to reproduce localization system errors on board of a self-driving vehicle and it is also useful to avoid that the \textit{active} agent follows the same route repetitively during the training phase.
\end{itemize}
Moreover, a further noise element is introduced by modifying the shape of the entry lanes (Fig.~\ref{fig:noise_envs}) every 1000 episodes in order to provide the \textit{active} agent different navigable spaces~(Fig.~\ref{fig:nav_space}) during the training phase.

\section{EXPERIMENTAL RESULTS}
\label{sec:exp_res}
\subsection{Training on Passive Agents}

\begin{table*}[b] \centering
\caption{Performances on the unseen scenario (Fig.~\ref{fig:synthetic_campus}) achieved by several approaches. Values are computed as the average of the three experiments with different traffic conditions (low, medium, high).}
\label{results}
\begin{center}
\begin{tabular}{|c|c|c|c|c|c|}
\hline
 & \textbf{Single\_env} & \textbf{Five\_envs} & \textbf{Five\_envs\&noise} & \textbf{Multi\_env} & \textbf{Multi\_env\&noise}\\
\hline
\textbf{Reaches \%} & 0.907 & 0.891 & 0.979 & 0.952 & 0.991 \\
\hline
\textbf{Crashes \%} & 0.093 & 0.109 & 0.021 & 0.048 & 0.009\\
\hline
\textbf{Total Steps} & 103.489 & 100.460 & 116.356 & 108.237 & 137.438\\
\hline
\end{tabular}
\end{center}
\end{table*}

As explained in Section~\ref{sec:passive_agents}, \textit{passive} agents should behave safely in all the roundabouts and therefore they are trained in all scenarios simultaneously in a \textit{Multi-environment Multi-agent} fashion. The maximum number of \textit{passive} agents learning and interacting among them simultaneously in each roundabout is reported in Table~\ref{max_passive_trained}. Moreover, considering the great diversity in the shape and size of roundabouts, we multiply the reward by a normalization factor, that is the ratio between the path lenght in which the rewards are computed ($l$) and the longest path measured among all the training scenarios ($l_{max}$).

At time $t$ the \textit{expected return} $R_t$ will be defined as follows:

\begin{equation}
R_t = \frac{l}{l_{max}} \sum_t^T r_t + \gamma r_{t+1} + \cdots + \gamma^{T-t}r_{T-t}
\end{equation}

\begin{figure}[h]
\centering
\includegraphics[width=0.8\linewidth]{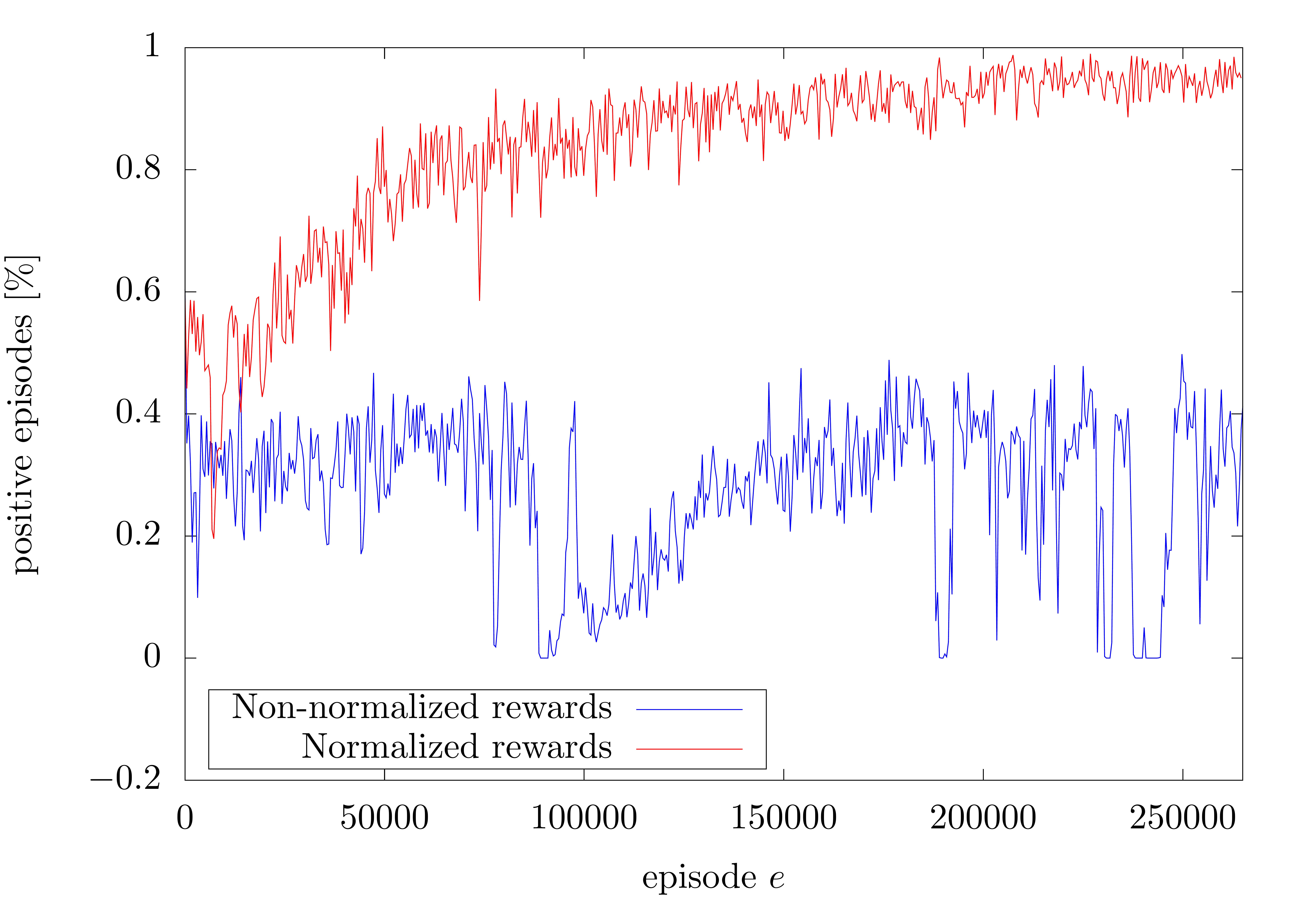}
\caption{Moving average of positive episodes ratio achieved during the training of \textit{passive agents} using normalized (red) and non-normalized (blue) rewards.}
\label{fig:passives_training}
\end{figure}

The learning curves in Fig.~\ref{fig:passives_training}, show the contribution of the normalization factor during the training of \textit{passive} agents.

\begin{table}[H] \centering
\caption{Maximum number of \textit{passive} vehicles trained in each scenario. In the following table \textit{Training roundabout 5} and \textit{Training roundabout 6} are the same \textit{validation} and \textit{test} scenarios used for \textit{active} agents respectively (Fig.~\ref{fig:synthetic_stadio}, Fig.~\ref{fig:synthetic_campus}) .}
\label{max_passive_trained}
\begin{center}
\begin{tabular}{|c|c|}
\hline
\textbf{Roundabouts} & \textbf{\#Passives}\\
\hline
Training roundabout 1 & 16\\
\hline
Training roundabout 2 & 8\\
\hline
Training roundabout 3 & 12\\
\hline
Training roundabout 4 & 16\\
\hline
Training roundabout 5 & 16\\
\hline
Training roundabout 6 & 24\\
\hline
\end{tabular}
\end{center}
\end{table}

\subsection{Insertion Maneuver Generalization}
\label{roundabout_test}

We evaluated the performance of the \textit{active} vehicles on the roundabout shown in Fig.~\ref{fig:synthetic_campus}, in order to compare our system with the results achieved in~\cite{iri}. For these agents the length of the episode does not change since it ends once the insertion in the roundabout is completed and for this reason there is no need to normalize their rewards as done for \textit{passives}.

As in~\cite{iri}, we performed three experiments, each one composed by 3000 episodes with different traffic conditions: \textit{low}, \textit{medium} and \textit{high} which correspond to a maximum number of \textit{passive} agents populating the \textit{test} roundabout to $10$, $15$ and $20$ respectively. The metrics used for these tests are: \textit{Reaches}, the percentage of episodes ended successfully, \textit{Crashes}, the percentage of episodes ended with a crash and \textit{Total Steps}, the average number of steps used to perform the whole experiment (3000 episodes). We replaced the metric \textit{Time-over} used in~\cite{iri} to calculate the percentage of the depletion of available time, with \textit{Total Steps} in order to compute the total amount of steps needed to end the episode, letting the agent performing the insertion without time limit. For this reason, we also performed a new test using the model described in~\cite{iri} (which we will call \textit{Single\_env}) following the new metrics and populating the \textit{test} roundabout with \textit{passive} agents trained as explained in Section~\ref{sec:passive_agents}.

Moreover, we analyzed the generalization performances on the \textit{test} roundabout with several models obtained by training the system with different configurations explained as follows:

\begin{itemize}
\item \textit{Five\_envs}: the system is trained in five environments (four roundabouts of Fig.~\ref{fig:synthetic_roundabouts} and the one in Fig.~\ref{fig:synthetic_stadio}) without the use of a \textit{validation} scenario (Fig.~\ref{fig:synthetic_stadio}) and noise injection;
\item \textit{Five\_envs\&noise}: we trained the module in five environments (four roundabouts of Fig.~\ref{fig:synthetic_roundabouts} and the one in Fig.~\ref{fig:synthetic_stadio}) using noise injection but without the \textit{validation} scenario (Fig.~\ref{fig:synthetic_stadio});
\item \textit{Multi\_env}: we use \textit{Multi-environment System} (Section~\ref{sec:multi-env_system}), using the \textit{validation} roundabout (Fig.~\ref{fig:synthetic_stadio}), but without noise injection (Section~\ref{sec:noise_inj});
\item \textit{Multi\_env\&noise}: the model is trained in \textit{Multi-environment System} (Section~\ref{sec:multi-env_system}) using both the \textit{validation} roundabout (Fig.~\ref{fig:synthetic_stadio}) and noise injection during training.
\end{itemize}

Table~\ref{results} shows the average percentages obtained with models described in the list above; we can notice from the results achieved by \textit{Five\_envs} model, that the mere addition of training environments does not avoid the risk of overfitting; indeed, the percentage of episodes ended successfully drops in the \textit{test} environment at a lower level than the one obtained with \textit{Single\_env}, even if it reaches good results ($>98\%$) in the scenarios in which such model is trained. Moreover, best results are achieved by those models trained using the \textit{validation} roundabout (Fig.~\ref{fig:synthetic_stadio}) or noise injection; in particular, \textit{Mulit\_env\&noise} considerably overcomes the results obtained in~\cite{iri} on the same \textit{test} roundabout (Fig.~\ref{fig:synthetic_campus}) and show the effects of the \textit{validation} roundabout (Fig.~\ref{fig:synthetic_stadio}) together with noise injection. Finally, we can notice that the introduction of unpredictable elements explained in Section~\ref{sec:noise_inj} are also useful to increase generalization performances in unseen scenarios even if they were designed to reduce the gap between synthetic and real data.

A further demonstration is given by the results illustrated in Table~\ref{validation_results} which represent the comparison between the performances achieved by the \textit{Multi\_env\&noise} and \textit{Single\_env} on the \textit{validation} roundabout (Fig.~\ref{fig:synthetic_stadio}). Both models do not use such scenario to update weights of their network parameters and for this reason the results in Table~\ref{validation_results} represent a good measure to evaluate generalization performances.

Due to a different size of this environment, the traffic used in this scenario is different from the one used on the \textit{test} roundabout: in this case, \textit{low}, \textit{medium} and \textit{high} correspond to the maximum number of $6, 12$ and $18$ \textit{passive} vehicles populating such scenario respectively.

\begin{table}[h] \centering
\caption{Performances on the \textit{validation} roundabout (Fig.~\ref{fig:synthetic_stadio}) achieved by \textit{Single\_env} and \textit{Multi\_env\&noise}. Values are computed as the average of the three experiments with different traffic conditions (low, medium, high).}
\label{validation_results}
\begin{center}
\begin{tabular}{|c|c|c|c|c|c|}
\hline
 & \textbf{Single\_env} & \textbf{Multi\_env\&noise} \\
\hline
\textbf{Reaches \%} & 0.920 & 0.994 \\
\hline
\textbf{Crashes \%} & 0.080 & 0.006 \\
\hline
\textbf{Total Steps} & 88.906 & 115.741\\
\hline
\end{tabular}
\end{center}
\end{table}

\subsection{Junction Insertions}

We performed a test on a different environment from roundabout scenarios, proving that the generalization achieved by our module achieves good results also for junction insertions. We chose this environment since the maneuver is similar to the one performed for roundabout insertions, but the navigable space as well as obstacle motions perceived by the \textit{active} vehicle are different from the roundabout case.

\begin{figure}[h]
  \centering
  \hspace{-2cm}
  \begin{subfigure}{.4\linewidth}
    \centering
    \includegraphics[width =\linewidth]{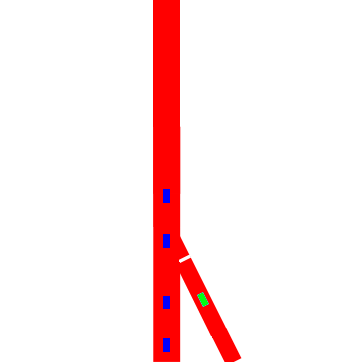}    
    \caption{Synthetic}
    \label{fig:synth_junction}
  \end{subfigure}
  \begin{subfigure}{.23\linewidth}
    \centering
    \includegraphics[width =\linewidth]{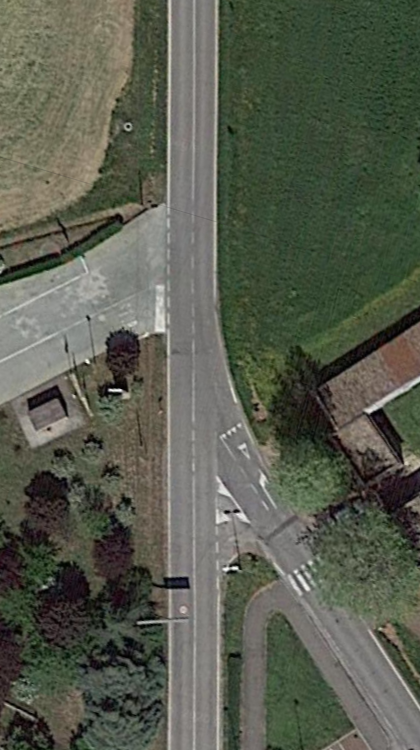}    
    \caption{Real}
    \label{fig:real_junction}
  \end{subfigure}  
  \caption{Real junction (b) of Parma and its synthetic representation (a).}
  \label{fig:junction}
\end{figure}

We test our system in the scenario illustated in Fig.~\ref{fig:junction}, performing three experiments, each one composed by 3000 episodes as in previous tests. In this scenario the different traffic conditions, \textit{low, medium} and \textit{high}, correspond to a maximum number of $2, 4$ and $6$ \textit{passive} vehicles populating the highway respectively, which are trained as in Section~\ref{sec:passive_agents}. Moreover, to evaluate the insertion module in different conditions, we reshape the entry lane at the beginning of each episode such that the junction angle could assume different values during tests.

Table~\ref{junction_results} shows the comparison between the results achieved by \textit{Multi\_env\&noise} with those obtained by the system developed in~\cite{iri} (\textit{Single\_env}), showing that our module features better generalization capabilities also in environments different from roundabout scenarios.

\begin{table}[H] \centering
\caption{Comparison between the results achieved by \textit{Multi\_env\&noise} and \textit{Single\_env} on the junction environment (Fig~\ref{fig:synth_junction}).}
\label{junction_results}
\begin{center}
\begin{tabular}{|c|c|c|}
\hline
 & \textbf{Single\_env} & \textbf{Multi\_env\&noise} \\
\hline
\textbf{Reaches \%} & 0.919 & 0.970 \\
\hline
\textbf{Crashes \%} & 0.081 & 0.030 \\
\hline
\textbf{Total Steps} & 97.011 & 128.182 \\
\hline
\end{tabular}
\end{center}
\end{table}

\subsection{Real-World Test}

\begin{figure}[h]
\centering
\includegraphics[width=0.7\linewidth]{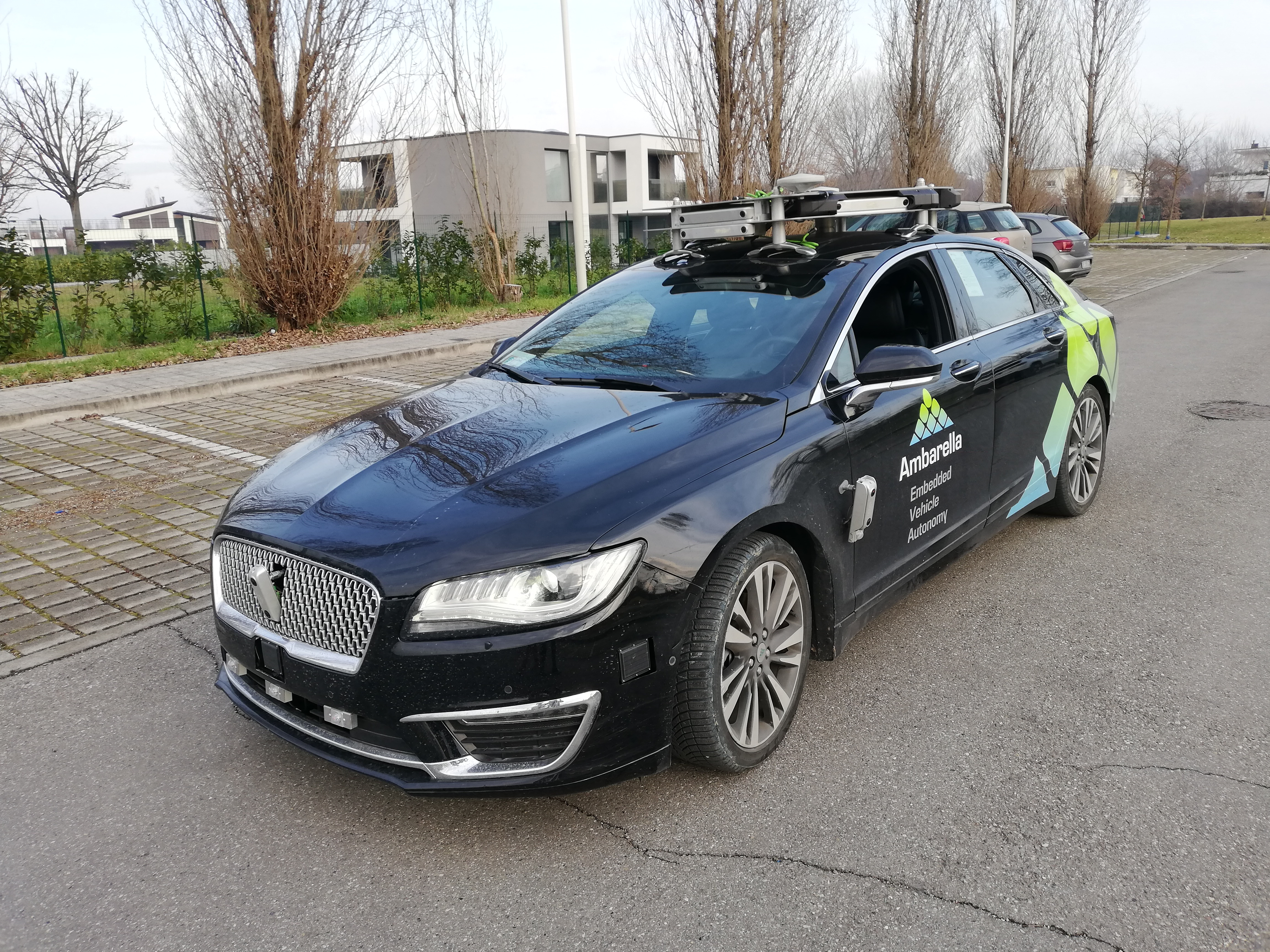}
\caption{Self-driving vehicle equipped with localization and perception systems used to perform tests in the real roundabout of Fig.~\ref{fig:real_parmamia}.}
\label{fig:lincoln}
\end{figure}

We performed real tests on board of a self-driving vehicle (Fig.~\ref{fig:lincoln}) in the roundabout illustrated in Fig.~\ref{fig:real_parmamia}, comparing the behavior of the best model achieved in this work (\textit{Multi\_env\&noise}) and the model developed in~\cite{iri} (\textit{Single\_env)}. We chose this scenario since the \textit{Single\_env} model is trained on the synthetic representation of such environment (Fig.~\ref{fig:synth_parmamia}) and because this roundabout is included in one of the few areas in which autonomous driving tests are allowed in Parma. We proved that the techniques developed in this work (Section~\ref{sec:techniques}) allow the vehicle to execute the insertion maneuver in such roundabout correctly; we noticed that it was infeasible to perform experiments using \textit{Single\_env} model since its behavior was unsure, with continuous changes of output due to uncertainty and noise introduced by localization and perception systems on board of the vehicle. In contrast, we proved that \textit{Multi\_env\&noise} model was more robust to these unpredictable elements and behaves correctly also in the real-world scenario used for this test, performing more than 100 immissions in the city traffic from the three entry lanes of the roundabout without errors. The following video (\url{https://youtu.be/QmgB0YH2BdQ}) shows some immissions performed by the self-driving vehicle equipped with the \textit{Multi\_env\&noise} model.

\section{CONCLUSION}

In this work we introduced some techniques useful to improve generalization both in unseen and real-world scenarios in the autonomous driving field, with particular care for the entering maneuver for roundabout insertions. We developed a \textit{Multi-environment System} consisting in four \textit{training} roundabouts (Fig.~\ref{fig:synthetic_roundabouts}) in which vehicles are trained simulataneously, and a \textit{validation} environment (Fig.~\ref{fig:synthetic_stadio}) used to select the best network parameters based on the results obtained on such scenario. Then, we developed techniques in order to reduce the gap between synthetic and real-world data, making the system more robust to noise that systems on board of a self-driving vehicle could introduce. Finally, we compared the performances between the best model achieved in this work (\textit{Multi\_env\&noise}) with the one obtained in~\cite{iri}~(\textit{Single\_env}) in the same unseen scenarios (Fig.~\ref{fig:synthetic_campus}, Fig.~\ref{fig:synthetic_stadio}) and in the real roundabout of Fig.~\ref{fig:real_parmamia}. We proved that \textit{Multi\_env\&noise} reaches better performances in both cases proving that the techniques developed in this work allow us to test our model on a self-driving vehicle, deploying our system in the real world even if it was fully trained in simulation.

\bibliography{root} 
\bibliographystyle{ieeetr}

\end{document}